# Inference with Idempotent Valuations

**Luis D. Hernández**
Dpto. Informática y Sistemas
Universidad de Murcia
30.071 - Espinardo - Murcia, Spain
e-mail:ldaniel@dif.um.es

**Serafín Moral**
Dpto. Ciencias de la Computación e I.A.
Universidad de Granada
18.071 - Granada - Spain
e-mail:smc@decsai.ugr.es

## Abstract

Valuation based systems verifying an idempotent property are studied. A partial order is defined between the valuations giving them a lattice structure. Then, two different strategies are introduced to represent valuations: as infimum of the most informative valuations or as supremum of the least informative ones. It is studied how to carry out computations with both representations in an efficient way. The particular cases of finite sets and convex polytopes are considered.

## 1 INTRODUCTION

Valuation Based Systems (VBS) were introduced by Shafer and Shenoy (Shafer and Shenoy 1988, Shafer 1991). This abstract framework to represent and calculate with valuations has been applied to different formalisms, such as Probability, Belief Functions (Shenoy and Shafer 1990), Convex Sets of Probabilities (Cano, Moral and Verdegay López 1993), Possibility Theory (Shenoy 1992), or Propositional Information Systems (Kohlas and Moral 1996).

In this paper, we study Valuation Based Systems in which an idempotent property is verified. This property allows to define a partial order in the set of valuations, which provides two methodologies to represent valuations: as infimum of very informative valuations and as supremum of very low informative ones.

Two particular cases are given to illustrate the theoretical developments: finite sets and convex sets of real numbers. It is shown as the general methods to represent valuations gives rise to the most common used techniques. For example, in the case of convex sets one corresponds to the representation by half-spaces and the other to the representation by extreme points.

In the computation part, each one of the basic operations -marginalization and combination- is very easy in one of the representations and difficult in the other. We concentrate in general results allowing to speed up the computations in the difficult cases.

In classical algorithms for valuations, the complexity of the computations is directly related to the size of the frame in which they are defined. With idempotent valuations it will be possible to handle valuations on a frame, with a complexity much lower than its size.

Kohlas and Stärk (1996) have introduced the concept of information algebra, which coincides essentially with an idempotent valuation system. They also define the associated order relation and show some of its properties. However, the focus of the paper is somewhat different. They presents this model as a general theory of information processing in computer science showing the relationships with Scott's information systems (Scott 1982) and tuple systems.

The paper is structured in the following way: Section 2 studies the basic of idempotent valuations. Section 3 considers the strategies for representing valuations. Section 4 is devoted to procedures for computing with valuations. Finally, the conclusions are in Section 5.

## 2 IDEMPOTENT VALUATION SYSTEMS

Assume that we have an n-dimensional variable $X = (X_1, \ldots, X_n)$ where each variable $X_i$ will take values on a set $\Omega_i = \Omega_{X_i}$. In these conditions, we will use the following notation:

- If $I \subseteq N = \{1, 2, \ldots, n\}$ then $X_I$ is the vector of variables $(X_i)_{i \in I}$ defined on $\Omega_I = \times_{i \in I} \Omega_i$. $\Omega_N$ is $\Omega_{\{1,\ldots,n\}}$.

- If $x \in \Omega_I$ and $J \subseteq I$ then $x^{\downarrow J}$ is the element of $\Omega_J$ obtained from $x$ but dropping the indexes of $I - J$. That is, $x_i^{\downarrow J} = x_i, \forall i \in J$.

Valuation Based Systems (VBS's) were introduced by Shafer and Shenoy (1988,1990). It is assumed the existence of a disjoint set $\mathcal{V}_I$ for each $I \subseteq \{1, \ldots, n\}$. The elements of $\mathcal{V}_I$ are called valuations on $\Omega_I$, and represent the different pieces of information we may have about $X_I$. The set of all the valuations, $\bigcup_{I \subseteq \{1,\ldots,n\}} \mathcal{V}_I$, will be denoted as $\mathcal{V}$.

For each valuation, $V$, $s(V)$ will denote the subset $I$, such that $V \in \mathcal{V}_I$.

There are two basic operations in valuation based systems (Shafer and Shenoy 1988, Shenoy and Shafer 1990):

- *Combination.-* It is a mapping $\otimes : \mathcal{V} \times \mathcal{V} \to \mathcal{V}$ such that $s(V_1 \otimes V_2) = s(V_1) \cup s(V_2)$.

- *Marginalization.-* Given $J \subseteq I$ and $V \in \mathcal{V}_I$, then the marginalization of $V$ to $J$ is a valuation $V^{\downarrow J} \in \mathcal{V}_J$.

Shenoy and Shafer (1988, 1990) consider the following axioms for these operations:

**Axiom 1.** $V_1 \otimes V_2 = V_2 \otimes V_1$, $(V_1 \otimes V_2) \otimes V_3 = V_1 \otimes (V_2 \otimes V_3)$

**Axiom 2.** If $I \subseteq J \subseteq K$ and $V \in \mathcal{V}_K$, then $(V^{\downarrow J})^{\downarrow I} = V^{\downarrow I}$.

**Axiom 3.** If $V_1 \in \mathcal{V}_I$, $V_2 \in \mathcal{V}_J$, then $(V_1 \otimes V_2)^{\downarrow I} = V_1 \otimes V_2^{\downarrow (J \cap I)}$.

Additionally, Cano, Delgado and Moral (1993) introduced the following two axioms, which will be also assumed for general VBS's.

**Axiom 4.** *Neutral Element.-* There is one and only one valuation $V_0$, defined on $\Omega_1 \times \ldots \times \Omega_n$, such that $\forall V \in \mathcal{V}^I$, $V_0^{\downarrow I} \otimes V = V$.

**Axiom 5.** *Contradiction.-* There is one and only one valuation $V_c$, defined on $\Omega_1 \times \ldots \times \Omega_n$, such that $\forall V \in \mathcal{V}$, $V_c \otimes V = V_c$.

An idempotent VBS is defined as a general VBS verifying the following axiom:

**Axiom 6.** $\forall V \in \mathcal{V}_J, \forall I \subseteq J$, $\quad V^{\downarrow I} \otimes V = V$

**Proposition 1** *If $\mathcal{V}$ is an idempotent VBS and $V \in \mathcal{V}$, such that $s(V) = I$, then $V^{\downarrow I} = V$.*

Proof

$$V^{\downarrow I} = (V \otimes V)^{\downarrow I} = V \otimes V^{\downarrow I} = V$$

**Example 1** *Let a valuation on $\Omega_I$ be a subset of $\Omega_I$. That is, $\mathcal{V}_I = 2^{\Omega_I}$. The combination of $V_1 \in \mathcal{V}_I$, $V_2 \in \mathcal{V}_J$ is defined as*

$$V_1 \otimes V_2 = \{x \in \Omega_{I \cup J} : x^{\downarrow I} \in V_1, x^{\downarrow J} \in V_2\} \quad (1)$$

*The marginalization of $V \in \mathcal{V}_I$ to $J \subseteq I$ is given by*

$$V^{\downarrow J} = \{x^{\downarrow J} : x \in V\} \quad (2)$$

*It is immediate to show that this is an idempotent VBS. The neutral valuation is $\Omega_1 \times \ldots \times \Omega_n$ and the contradiction is the empty subset in $\Omega_1 \times \ldots \times \Omega_n$.*

**Example 2** *Consider that each set $\Omega_i$ is the set of real numbers, $\mathbb{R}_i$. A valuation on $\mathbb{R}_I$ is a convex set of points from $\mathbb{R}_I$, i.e. a subset $V \subseteq \mathbb{R}_I$ such that if $x, y \in V$, and $\alpha \in [0, 1]$, then $\alpha x + (1 - \alpha) y \in V$.*

*The combination of $V_1 \in \mathbb{R}_I$, $V_2 \in \mathbb{R}_J$ is defined as*

$$V_1 \otimes V_2 = \{x \in \mathbb{R}_{I \cup J} : x^{\downarrow I} \in V_1, x^{\downarrow J} \in V_2\} \quad (3)$$

*The marginalization of $V \in \mathbb{R}_I$ to $J \subseteq I$ is given by*

$$V^{\downarrow J} = \{x^{\downarrow J} : x \in V\} \quad (4)$$

*It is immediate to show that this is an idempotent VBS. Definitions are quite similar to the example above, but as we shall show later in the paper, there are meaningful differences between these two systems.*

Below we define a partial order relation in idempotent VBS's, which is very useful to study them.

**Definition 1** *We will say that $V_1$ is less informative than $V_2$ ($V_1 \preceq V_2$), iff $V_1 \otimes V_2 = V_2$.*

**Example 3** *In both examples above (ordinary subsets and convex sets), $V_1 \preceq V_2$ if and only if $V_2 \subseteq V_1$.*

Some of the properties of the following proposition (2 and 5) have been also shown by Kohlas and Stärk (1996).

**Proposition 2** *The relation $\preceq$ verifies the following properties:*

1. *It is a (reflexive) partial order.*

2. *If $V_1 \in \mathcal{V}_I, V_2 \in \mathcal{V}_J$, $V_1 \preceq V_2$, and $K \subseteq J$, then $V_1^{\downarrow K} \preceq V_2^{\downarrow K}$*

3. *Sup $\mathcal{V} = V_c$*

4. *Inf $\mathcal{V}_I = V_0^{\downarrow I}$*

5. *$V_1 \otimes V_2 =$ Sup $\{V_1, V_2\}$*

6. $V^{\downarrow J} = \text{Sup } \{V' \in \mathcal{V}_J : V' \preceq V\}$

Proof

*Properties 1-5 are easily obtained from valuation axioms (including indempotence).*

*To show 6, consider $V' \in \mathcal{V}_J$, such that $V' \preceq V$, then following Axiom 3,*

$$V^{\downarrow J} \otimes V' = (V \otimes V')^{\downarrow J} \quad (5)$$

*Since $V' \preceq V$, then $V \otimes V' = V$, and we can write*

$$V^{\downarrow J} \otimes V' = V^{\downarrow J} \quad (6)$$

*As a consequence: $V' \preceq V^{\downarrow J}$ and $V^{\downarrow J}$ is an upper bound of $\{V' \in \mathcal{V}_J : V' \preceq V\}$.*

*To prove that it is the supremum, consider that $V'' \in \mathcal{V}_J$ is an upper bound of the set above. As $V^{\downarrow J} \preceq V$, then $V^{\downarrow J} \preceq V''$, and therefore $V^{\downarrow J}$ is the minimum of the upper bounds (the supremum).*

As a consequence of property 5 above the supremum of 2 valuations always exists and it is equal to its combination.

**Definition 2** *If $V \in \mathcal{V}_I$ and $I \subseteq J$, then the extension of $V$ to $J$ is defined as the valuation $V^{\uparrow J} = V \otimes V_0^{\downarrow J}$.*

It is immediate to verify that if $V \in \mathcal{V}_I$ and $J \subseteq I$ then $(V^{\downarrow J})^{\uparrow I} \preceq V$. On the other hand, if $I \subseteq K$ then $(V^{\uparrow K})^{\downarrow I} = V$.

**Definition 3** *If $V_1 \in \mathcal{V}_I$ and $V_2 \in \mathcal{V}_2$, then $V_1 \oplus V_2$ is equal to $\text{Inf } \{V_1, V_2\} = \text{Inf } \{V_1^{\uparrow I \cup J}, V_2^{\uparrow I \cup J}\}$ under $\preceq$ relation.*

There is no guarantee that $V_1 \oplus V_2$ exists for every pair of valuations.

**Example 4** *In the example in which valuations are subsets from $\Omega_I$, $V_1 \oplus V_2$ always exists and it is equal to $V_1^{\uparrow I \cup J} \cup V_2^{\uparrow I \cup J}$. First, the two sets are extended to a common frame and then the union is calculated.*

*In the case of convex sets, $V_1 \oplus V_2$ is the convex hull (the minimum convex set containing it) of the union of the extensions: $CH(V_1^{\uparrow I \cup J} \cup V_2^{\uparrow I \cup J})$.*

**Proposition 3** *If $V_1 \in \mathcal{V}_I, V_2 \in \mathcal{V}_J$ are such that $(V_1 \oplus V_2)$ exists, and $K \subseteq (I \cap J)$, then*

$$(V_1 \oplus V_2)^{\downarrow K} = (V_1^{\downarrow K} \oplus V_1^{\downarrow K}) \quad (7)$$

Proof

*We have that $V_1^{\downarrow K} \preceq V_1$ and $V_2^{\downarrow K} \preceq V_2$. Thus, $V_1^{\downarrow K} \oplus V_2^{\downarrow K} \preceq V_1 \oplus V_2$ and taking into account that $(V_1 \oplus V_2)^{\downarrow K} = \text{Sup } \{V : V \in \mathcal{V}_K, V \preceq V_1 \oplus V_2\}$, we obtain that*

$$(V_1^{\downarrow K} \oplus V_2^{\downarrow K}) \preceq (V_1 \oplus V_2)^{\downarrow K} \quad (8)$$

*On the other hand, $(V_1 \oplus V_2) \preceq V_1^{\uparrow I \cup J}$ and therefore $(V_1 \oplus V_2)^{\downarrow K} \preceq (V_1^{\uparrow I \cup J})^{\downarrow K} = V_1^{\downarrow K}$. Analogously, we can obtain $(V_1 \oplus V_2)^{\downarrow K} \preceq V_2^{\downarrow K}$. As a consequence,*

$$(V_1 \oplus V_2)^{\downarrow K} \preceq (V_1^{\downarrow K} \oplus V_1^{\downarrow K}) \quad (9)$$

*As the relation $\preceq$ is antisymmetric, equations (8) and (9) proves the proposition.*

In the following we define the combination of a possibly infinite set of valuations. As for finite sets the combination of all the valuations is the supremum (a consequence of property 5 in Proposition 2), then the following definition is consistent with the finite case.

**Definition 4** *Let $H \subseteq \mathcal{V}$, the combination of the valuations in $H$, $\bigotimes H$, is equal to*

$$\text{Sup } \{V : V \in H\} \quad (10)$$

When $H$ is finite this supremum exists and it is the repeated combination of all the elements of $H$. However, in general, there is no guarantee that this supremum exists.

**Definition 5** *Let $H \subseteq \mathcal{V}$, the disjunction of the valuations in $H$, $\bigoplus H$, is equal to*

$$\text{Inf } \{V^{\uparrow I} : V \in H\} \quad (11)$$

*where $I = \bigcup_{V \in H} s(H)$.*

$\bigoplus H$ is the infimum of the extension of the valuations in set $H$ to a common frame.

**Definition 6** *An idempotent VBS is said to be complete if and only if $\bigotimes H$ exists for every $H \subseteq \mathcal{V}$.*

**Proposition 4** *If $\mathcal{V}$ is an idempotent and complete VBS then $\bigoplus H$ exists for every $H \subseteq \mathcal{V}$.*

Proof

*If $H \subseteq \mathcal{V}$, let $I = \bigcup_{V \in H} s(V)$ and $L_H$ the set of the lower bounds of the extensions of valuations in $H$ to $I$, under $\preceq$ relation, i.e.,* $L_H = \{V : V \preceq V'^{\uparrow I}, \forall V' \in H\}$. $L_H$ is not empty, because $V_0^{\downarrow I}$ belongs always to it.

As $\mathcal{V}$ is complete $\bigotimes L_H$ exists. Let us show that it is the infimum $\bigoplus H$. First, let us show that it is a lower bound. If $V' \in H$, then $\forall V \in L_H$, $V \preceq V'^{\uparrow I}$. So, $(\bigotimes L_H) \preceq V'^{\uparrow I}$. An then $(\bigotimes H)$ is a lower bound of the extensions of valuations in $H$ to $Y$.

On the other side, If $V^*$ is a lower bound of the extensions of valuations in $H$, then $V^* \in L_H$ and therefore $V^* \preceq (\bigotimes L_H)$. So $(\bigotimes L_H)$ is the maximum of lower bounds, i.e., the supremum.

## 3  REPRESENTATION SYSTEMS

In the following, we express a valuation as composition of simpler valuations. We assume that $\mathcal{V}$ is an idempotent VBS.

**Definition 7** *A subset $\mathcal{V}_* \subseteq \mathcal{V}$ will be called a lower representation system (LRS) iff $\forall I, V_0^{\downarrow I} \notin \mathcal{V}_*$, and*

$$\forall V \in \mathcal{V}_I, \ \exists H \subseteq \mathcal{V}_*, \text{ such that } V = \otimes H \quad (12)$$

*where $H$ is finite.*

That is, each valuation $V$ can be expressed as a combination of a finite set of valuations $V' \in \mathcal{V}_*$. These valuations, $V'$, are defined on frames $s(V') \subseteq s(V)$.

The representation of a valuation in a lower representation system is not unique.

**Definition 8** *$H \subseteq \mathcal{V}_*$ is said to be a minimal representation of $V$ if and only if $V = \bigotimes H$ and if $H' \subseteq H$ is such that $V = \bigotimes H'$, then $H' = H$.*

The following proposition is immediate.

**Proposition 5** *$H$ is minimal if and only if there is no $V' \in H$ such that $(\bigotimes H) = (\bigotimes H - \{V'\})$.*

It is evident that $\mathcal{V}$ is a representation system because each valuation $V$ can be expressed as a combination of itself: $V = \otimes\{V\}$. But this will be a very useless one. We want represent valuations with simpler valuations. This will be achieved with the so-called basic representations systems.

**Definition 9** *A set of valuations $\mathcal{V}_{**}$ is weakly included in the set of valuations $\mathcal{V}_*$, $\mathcal{V}_{**} \sqsubseteq \mathcal{V}_*$, iff*

$$\forall V \in \mathcal{V}_{**} \ \exists V' \in \mathcal{V}_* \text{ verifying } V \preceq V' \text{ (i.e. } V' = V \otimes V') \quad (13)$$

**Definition 10** *A basic lower representation system (BLRS) $\mathcal{V}_*$ on $\mathcal{V}$ is a lower representation system which verifies:*

$$\begin{aligned}&\not\exists \mathcal{V}_{**} \subseteq \mathcal{V}, \text{ representation system,}\\ &\text{such as } \mathcal{V}_{**} \sqsubseteq \mathcal{V}_* \text{ and } \mathcal{V}_* \neq \mathcal{V}_{**}\end{aligned} \quad (14)$$

**Theorem 1** *If $\mathcal{V}_*$ is a BLRS and $H$ is a representation of $V \in \mathcal{V}_*$ then $V \in H$.*

### Proof

Let $V \in \mathcal{V}_*$, and $H$ a representation of $V$ in $\mathcal{V}_*$. If $V \notin H$, let us consider $\mathcal{V}_{**} = \mathcal{V}^* - \{V\}$.

Then $\mathcal{V}_{**}$ is a lower representation system: if a valuation $V'$ has as representation $H'$ in $\mathcal{V}_*$, then if $V \notin H'$, the same $H'$ is a representation of $V'$ in $\mathcal{V}_{**}$. If $V \in H'$, then a representation of $V'$ in $\mathcal{V}_{**}$ can be obtained by considering $H'' = (H' - \{V\}) \cup H$.

It is also immediate to verify that $\mathcal{V}_{**} \sqsubseteq \mathcal{V}_*$, and this is in contradiction with the fact that $\mathcal{V}_*$ is a basic lower representation system.

**Theorem 2** *If $\mathcal{V}_*$ is a BLRS and $\mathcal{V}_{**}$ is a LRS, then $\mathcal{V}_* \subseteq \mathcal{V}_{**}$*

### Proof

If $V \in \mathcal{V}_*$, since $\mathcal{V}_{**}$ is a LRS, then there exists $H \subseteq \mathcal{V}_{**}$, such that $V = \bigotimes H$.

Now for each $V' \in H$, let $H_{V'} \subseteq \mathcal{V}_*$ be a representation of $V'$ in BLRS $\mathcal{V}_*$.

If we consider $H' = \bigcup_{V' \in H} H_{V'}$. Then, it is easy to show that $H'$ is a representation of $V$ in $\mathcal{V}_*$, and by Theorem 1, we can deduce $V \in H'$. Consider $V'$ a valuation in $H$ such that $V \in H_{V'}$.

Since $V' = \bigotimes H_{V'}$, then $V \preceq V'$.

Since $V = \bigotimes H$ and $V' \in H$, then $V' \preceq V$.

As the inclusion is anti-symmetrical we have that $V = V'$ and as $V' \in \mathcal{V}_{**}$, we have that $V \in \mathcal{V}_{**}$.

**Corollary 1** *It there exists a basic lower representation system in $\mathcal{V}$, then it is unique.*

**Corollary 2** *If $\mathcal{V}_*$ is a BLRS then*

$$\mathcal{V}_* = \bigcap \{\mathcal{V}_{**} : \mathcal{V}_{**} \text{ is a LRS}\} \quad (15)$$

**Example 5** *For the VBS of subsets, and when $\Omega_i$ is finite for every $i \in \{1, \ldots, n\}$, a basic lower representation system can be obtained by considering the least* *informative subsets in each frame, $\Omega_I$, other than the neutral element. They are given by the complementary of one element $x_I \in \Omega_I$: $\neg x_I$.*

*A representation of a set $A \subseteq \Omega_I$ can be given by the set $H = \{\neg x_I : x_I \notin A\}$, i.e., by enumerating all the elements not in A.*

*In general, a valuation $A \subseteq \Omega_I$ is more efficiently represented by using basic valuations on frames $\Omega_J$, where $J \subset I$. These valuations are more informative. Assume, for example, that $\Omega_i = \{0,1\}, \quad \forall i = 1, \ldots, n$, and the subset given by:*

$$A = \{x \in \Omega_N : (x_n = 1 \Leftrightarrow x_i = 1, \forall i \leq n-1)\} \quad (16)$$

*If we use only valuations in $\Omega_N$ to define this set, we need to list the $2^{n-1}$ elements of the complementary of A. However, this set can be efficiently represented by valuations $H = \{V_1, \ldots, V_{n-1}, T\}$, where,*

$$V_i \equiv \neg(0,1) \subseteq \Omega_{i,n}, \quad i = 1, \ldots, n-1 \quad (17)$$

$$T \equiv \neg(1, \ldots, 1, 0) \subseteq \Omega_N \quad (18)$$

*The order of the representation is $O(n)$.*

*If we interpret the elements of $\Omega_i$ as the true values of a propositional symbol $p_i$, then a basic valuation corresponds to a clause. A clause as $p_1 \vee \neg p_2 \vee p_3$ says that the assignation of true values: $(0,1,0)$ is impossible. So it can be represented by $\neg(0,1,0) \in \Omega_{\{1,2,3\}}$. In general, the representation of a formula is obtained by expressing it in conjunctive normal form.*

**Example 6** *In the case of general convex sets in $\mathbb{R}_I$, there is not a basic lower representation system. We have to make a restriction and to consider only polyhedral convex sets. In this case, a basic representation system is given by the set of all the half-spaces in each $\mathbb{R}_I$ (a half-space is the portion of the space lying on one side of a hyperplane). Each basic valuation in $\mathbb{R}_I$ is given by a linear restriction:*

$$\sum_{x_I \in \mathbb{R}_I} \alpha_{x_I} . x_I \leq \beta \quad (19)$$

*In this case, we do not obtain a great reduction by considering valuations defined on smaller frames. However, it is generally admitted that giving a polyhedral by a minimal set of half-spaces is an efficient representation.*

Now, we consider upper representation systems, in which a valuation is given by taking the infimum of a finite set of valuations, i.e., we use more informative valuations to express a given valuation. Definitions and results are similar to the case of lower representation systems. So no proof of the results is given.

**Definition 11** *A subset $\mathcal{V}^* \subseteq \mathcal{V}$ will be called an* upper representation system *(URS) iff $\forall I, V_c^{\downarrow I} \notin \mathcal{V}^*$, and*

$$\forall V \in \mathcal{V}_I, \quad \exists H \subseteq \mathcal{V}^* \cap \mathcal{V}_I, \quad \text{such that } V = \oplus H \quad (20)$$

*where H is finite.*

The set of all possible upper representation systems of $\mathcal{V}$ will be denoted by $URS(\mathcal{V})$.

**Definition 12** *$H \subseteq \mathcal{V}^*$ is said to be a minimal representation of V if and only if $V = \bigoplus H$ and verifying that if $H' \subseteq H$ is such that $V = \bigoplus H'$, then $H' = H$.*

The following proposition is immediate.

**Proposition 6** *H is minimal if and only if there is no $V' \in H$ such that $(\bigoplus H) = (\bigoplus H - \{V\})$.*

**Definition 13** *A set of valuations $\mathcal{V}^{**}$ weakly includes the set of valuations $\mathcal{V}^*$, $\mathcal{V}^{**} \sqsupseteq \mathcal{V}^*$, iff*

$$\forall V \in \mathcal{V}^{**} \quad \exists V' \in \mathcal{V}^* \text{ verifying } V' \preceq V \quad (21)$$

**Definition 14** *A basic upper representation system (BURS) $\mathcal{V}^*$ on $\mathcal{V}$ is an upper representation system which verifies:*

$$\begin{array}{c}\nexists \mathcal{V}^{**} \subseteq \mathcal{V}, \text{ upper representation system,} \\ \text{such as } \mathcal{V}^{**} \sqsupseteq \mathcal{V}^* \text{ and } \mathcal{V}^* \neq \mathcal{V}^{**}\end{array} \quad (22)$$

**Theorem 3** *If $\mathcal{V}^*$ is a BURS and H is a representation of $V \in \mathcal{V}^*$ then $V \in H$.*

**Theorem 4** *If $\mathcal{V}^*$ is a BURS and $\mathcal{V}^{**}$ is an URS, then $\mathcal{V}^* \subseteq \mathcal{V}^{**}$*

**Corollary 3** *It there exists a basic upper representation system in $\mathcal{V}$, then it is unique.*

**Corollary 4** *If $\mathcal{V}^*$ is a BURS then*

$$\mathcal{V}^* = \bigcap \{\mathcal{V}^{**} : \mathcal{V}^{**} \text{ is an URS}\} \quad (23)$$

**Example 7** *For the VBS of subsets and being $\Omega_i$ finite for each $i = 1, \ldots, n$, and upper representation system can be obtained by considering the most informative valuations in each $\Omega_i$: the elements, $x_I \in \Omega_I$. So, a set is represented by listing its elements.*

*Again, more efficiency is obtained if we consider basic sets in smaller frames $\Omega_J, J \subset I$. Consider the same subset A of Example 5. This set may be represented by $\bigoplus H'$, where $H' = \{V'_1, \ldots, V'_{n-1}, T'\}$ and*

$$V_i' \equiv (0,0) \subseteq \Omega_{\{i,n\}}, \quad i = 1, \ldots, n-1 \quad (24)$$

$$T' \equiv (1, \ldots, 1) \subseteq \Omega_N \quad (25)$$

*Again the representation is $O(n)$. Anyway, there are cases which can be efficiently represented in one of the basic systems (lower or upper) and not in the other, and cases which can not be efficiently represented in any of them.*

*If each $\Omega_i$ is the set of true values of $p_i$, then a basic upper valuation is a conjunction of atomic formulas: $\neg p_1 \wedge \neg p_2 \wedge p_3$ corresponds to the basic set $(0,0,1) \in \Omega_{\{1,2,3\}}$. The representation of a general formula is obtained by expressing it in disjunctive normal form.*

**Example 8** *The most informative convex sets in $\mathbb{R}_I$ are the points: $x_I \in \mathbb{R}_I$. However general convex sets, or even polyhedral, can not be represented as the convex hull (the infimum) of a finite set of points. We must do some additional restriction in the type of convex sets we want to consider. One possibility is to consider the polytopes in which each component $x_i \in [0,1]$, i.e., each convex set $V \in \mathbb{R}_I$, is included in the hypercube $\times_{i \in I}[0,1] = [0,1]_I$, which is now the neutral element.*

*Here, considering points in lower dimensions has little interest: there is not to much gain in efficiency. However, representing a convex set by its extreme points is considered as the alternative representation to use half-spaces (Preparata and Shamos 1985).*

*The transformation between both representations has achieved a great deal of attention in the field of Computation Geometry. A survey of algorithms to find the extreme points from the half-spaces can be found in Mattheis and Rubin (1977). The half-spaces of a polytope defined as a set of extreme points are calculated by applying the so-called convex hull algorithms (Preparata and Shamos 1985).*

## 4 COMPUTING WITH REPRESENTATIONS

In general, the problem of reasoning with valuations can be stated in the following way: we have a finite set of valuations $R = \{V_1, \ldots, V_m\}$, where each $V_i$ is defined on a frame $I_i$. We are interested in calculating (Shafer and Shenoy 1988):

$$R_j = \left(\bigotimes R\right)^{\downarrow\{j\}} = (V_1 \otimes \ldots \otimes V_m)^{\downarrow\{j\}} \quad (26)$$

for a particular $j \in \{1, \ldots, n\}$.

This is done by the so-called deletion algorithm (Shafer and Shenoy 1988, Shenoy and Shafer 1990). This algorithm proceeds by transforming the set $R = \{V_1, \ldots, V_m\}$ according to the following basic step (*Deleting index $k$*):

- Let $k$ be an index, $k \neq j$. Consider $K = \{V_i \in R : k \in s(V_i)\}$ and $L = s(\bigotimes K) - \{k\}$. Transform $R$ into the set

$$R - K \cup \{(\otimes K)^{\downarrow L}\} \quad (27)$$

In our case, as we want to use only basic valuations we replace $K$ by a subset of basic valuations representing $(\otimes K)^{\downarrow L}$.

This step is repeated until all the valuations are defined on frame $\{j\}$. Then the solution, $R_j$, is represented by all the valuations in $R$.

To make notation more expressive, $V^{\downarrow s(V) - \{k\}}$ will be denoted as $V^{-k}$. So $(\otimes K)^{\downarrow L}$ will be expressed as $(\otimes K)^{-k}$.

In this section we study how this basic step can be carried out with valuations represented in a BLRS and in a BURS.

### 4.1 USING A LOWER REPRESENTATION SYSTEM

First, let us consider that each $V_i$ is given by $\bigotimes H_i$, where $H_i \subseteq \mathcal{V}_*$ and $\mathcal{V}_*$ is a BLRS.

In this case the combination $\bigotimes K$ is very easy: $\bigotimes K = \bigotimes H_K$, where $H_K = \bigcup_{V_i \in K} H_i$. That is, a representation of the combination of the valuations in $K$ is obtained by calculating the union of the representations of each one of the valuations in $K$.

Next we have to calculate $(\bigotimes H_K)^{-k}$. This marginalization is not immediate, but before doing it, it is convenient to remove from $H_K$ the subsumed valuations: a valuation $V \in H_K$ is subsumed by $V' \in H_K$ if and only if $V \preceq V'$. It is clear that we can remove subsumed valuations from $H_K$ without changing the result of the combination. Removing valuations can make future calculations more efficient.

Let $H_K'$ and $H_K''$ be defined in the following way:

$$H_K' = \{V \in H_K : k \in s(V)\}, \quad H_K'' = H_K - H_K' \quad (28)$$

It is clear that,

$$(\otimes H_K)^{-k} = (\otimes H_K'') \otimes (\otimes H_K')^{-k} \quad (29)$$

So basic valuations, $H_K''$, which are not defined for the variable to delete, $k$, may be set aside and to consider only valuations in $H'_K$. Afterwards, $H''_K$ is added to the representation of $(\otimes H'_K)^{-k}$.

In general, to calculate $(\otimes H'_K)^{-k}$ it is not necessary to combine all the valuations in $H'_K$ before carrying out the marginalization. In most of the cases, it is enough to make simpler combinations. The basis to do it will be to try to verify the condition in the following definition.

**Definition 15** *In a BLRS $\mathcal{V}_*$ the deletion of index $k$ is of dimension $l$ if and only if for every set $H \subseteq \mathcal{V}_*$, such that $k \in s(V), \forall V \in H$, the following condition is verified:*

$$(\otimes H)^{-k} = \bigotimes (\{(\otimes H')^{-k} : H' \subseteq H, |H'| = l\}) \tag{30}$$

*where $|H'|$ stands for the cardinal of $H'$.*

So if the deletion of $k$ is of dimension $l$, we have not to combine all the valuations in $H'_K$. We have to consider all the subsets, $H'$, of $H'_K$ with $l$ elements and then to calculate $(\otimes H')^{-k}$. The representation of the marginalization is the union of the representations of all the sets $(\otimes H')^{-k}$. The efficiency of the resulting procedure is based on using strategies to select only those subsets $H' \subseteq H'_K$ such that $H'^{-k}$ is not the neutral element.

Each time we calculate new valuations in $(\otimes H'_K)^{-k}$, we can check whether there are valuations in $H'_K$ subsumed by them: they can be removed from $H'_K$, decreasing the number of future subsets $H'$.

**Example 9** *In the VBS of subsets where each $\Omega_i$ is finite, the deletion of index $k$ is of dimension $|\Omega_k|$. Assume that we have a set of basic valuations $H = \{\neg x^1, \ldots, \neg x^m\}$ where each $x^j$ has index $k$. Then the deletion of $k$ can be carried out in the following way:*

1. *Determine the subsets $H' \subseteq H$, with $|\Omega_k|$ elements and such that*

$$\forall \neg x^{i_1}, \neg x^{i_2} \in H', \ x^{i_1}_k \neq x^{i_2}_k \tag{31}$$

*If $\neg x^{i_1}, \neg x^{i_2} \in H', i \in s(\neg x^{i_1}) \cap s(\neg x^{i_2})$ and $(i \neq k)$ then $x^{i_1}_i = x^{i_2}_i$*

$$\tag{32}$$

2. *For each one of these subsets $H'$, let $\neg x^{H'}$ the basic valuation defined on $s(\otimes H') - \{k\}$, and given by $x^{H'}_i = x^j_i$, where $\neg x^j \in H'$.*

$(\otimes H)^{-k}$ *is represented by the set of all the $\neg x^{H'}$ obtained with rules above.*

*When each $\Omega_i$ has the true values of a proposition $p_i$ as elements, then Step 1 above, selects a pair of clauses:* *one with $p_k$ and the other with $\neg p_k$ and such that its resolution by $p_k$ is not trivial (they do not have another pair of opposite literals). Step 2 does the resolution of the two clauses, resulting on a deletion of $p_k$. What it is done is an ordered resolution of all the clauses in $H$ to find all the consequences not containing $p_k$ (Kohlas and Moral 1996).*

*Above methodology is a generalization of resolution for arbitrary finite sets.*

*Assume now that we have the set $A$ of Example 5 represented by valuations $\{V_1, \ldots, V_{n-1}, T\}$ and that we have observed that $p_2$ is false, i.e. the valuation $O_2$ given by $\neg(1) \subseteq \Omega_2$. In order to calculate $\{A \otimes O_2\}^{\downarrow\{n\}}$, the deletion algorithm proceeds as follows:*

- *To delete $k = 1$, there are two basic valuations defined for this index: $T$ and $V_1$, where*

$$T \equiv \neg(1, \ldots, 1, 0) \subseteq \Omega_N, \quad V_1 \equiv \neg(0, 1) \subseteq \Omega_{1,n}$$

*By applying our deletion procedure, these valuations are removed and no new basic valuation is added to $H$.*

- *For $k = 3, \ldots, n-1$, we proceed by removing all valuations, $V_k$, without adding new valuations,*

- *For $k = 2$, we have the valuations $O_2$ and $V_2$. Its resolution produces as result $\neg(1) \subseteq \Omega_n$, i.e., $\neg p_n$.*

*In this moment, with all the valuations defined in $\Omega_n$ we stop and give the desired result: $\neg p_n$. The complexity of the computations has been much more lower than the size of the frame $\Omega_N$. In fact, it has been or order $O(n)$.*

**Example 10** *In the case of polyhedral in $\mathbb{R}_I$, the deletion of index $k$ is always of dimension 2. Assume that we have a subset $H = \{V_1, \ldots, V_n\}$ where each $V_i$ is a half-space in $\mathbb{R}_{I_i}$ and $k \in I_i$. The deletion of index $k$ can be done as follows. Let $L_i$ be the linear restriction defining half-space $V_i$,*

$$L_i \equiv \sum_{j \in I_i} \alpha^j_i x_j \leq \beta_i \tag{33}$$

*Let $\mathcal{L}^+$ be the set of linear restrictions, $L_i$, for which $\alpha^k_i > 0$, $\mathcal{L}^-$ be the set of restrictions for which $\alpha^k_i < 0$, and $\mathcal{L}^0$ the set of linear restrictions for which $\alpha^k_i = 0$.*

*For each pair $L_i \in \mathcal{L}^+$ and $L_j \in \mathcal{L}^-$ consider the restriction $L_{i,j} \equiv -\alpha^k_j.L_i + \alpha^k_i.L_j$, which is given by*

$$\sum_{l \in I_i \cap I_j - \{k\}} (-\alpha_j^k \alpha_i^l + \alpha_i^k \alpha_j^l) x_l +$$
$$\sum_{l \in I_i - I_j} (-\alpha_j^k \alpha_i^l) x_l + \quad (34)$$
$$\sum_{l \in I_j - I_i} (\alpha_i^k \alpha_j^l) x_l \leq -\alpha_j^k \beta_i + \alpha_i^k \beta_j$$

Let $\mathcal{L}^{-k} = \{L_{i,j} : L_i \in \mathcal{L}^+, L_j \in \mathcal{L}^-\} \cup \mathcal{L}^0$.

Then $H^{-k}$ is the valuation given by half-spaces defined by restrictions in $\mathcal{L}^{-k}$.

By analogy, we call the restriction $L_{i,j}$, the resolution of $L_i$ and $L_j$.

This method of calculating marginalization of convex sets is used in Preparata and Shamos (1985). We do not know how this simple technique compares with another more complicated procedures such as quantifier elimination methods (Lassez and Lassez 1992).

## 4.2 USING AN UPPER REPRESENTATION SYSTEM

Now for each valuation, $V_i$, there is a subset of basic upper valuations, $H_i$, such that $V_i = \bigoplus H_i$. By Proposition 3, the easy step in the calculation of $(\bigotimes K)^{\downarrow L}$ is the marginalization and the difficulty is in the combination $(\bigotimes K)$. To simplify this exposition, let us consider the case of combining two valuations, $V_1$ and $V_2$, given by sets, $H_1$ and $H_2$. The extension to a finite number of valuations is immediate.

We want to calculate:

$$(\oplus H_1) \otimes (\oplus H_2) \quad (35)$$

Now a previous step is to remove from each $H_i$ the valuations subsuming another valuation in $H_i$.

The basis of the computations will be to verify the following definition.

**Definition 16** A set of basic upper valuations, $H$, is said to be of dimension $l$ if and only if $\forall V \in \mathcal{V}^*$, if $\oplus H \preceq V$, then there is a subset $R \subseteq H$ with $|R| = l+1$, and $\oplus R \preceq V$.

If $H_1$ and $H_2$ are of dimensions $l_1$ and $l_2$, then:

$$(\oplus H_1) \otimes (\oplus H_2) =$$
$$\oplus\{(\oplus R_1) \otimes (\oplus R_2) : R_i \subseteq H_i, |R_i| = l_i, i = 1, 2\} \quad (36)$$

This is a kind of generalized distributive property which transform the problem in doing more combinations but with much simpler valuations.

**Example 11** In the case of valuations which are finite subsets of $\Omega_I$, then every set $H$ is of dimension 0. So we obtain the distributive property:

$$(\oplus H_1) \otimes (\oplus H_2) = \oplus\{V_1 \otimes V_2 : V_1 \in H_1, V_2 \in H_2\} \quad (37)$$

More efficiency is obtained when we only select non-inconsistent valuations $V_1$ and $V_2$. Since a valuation is an element, $V_i = x_{I_i}$, then this condition is verified when $x_{I_1}^{\downarrow I_1 \cap I_2} = x_{I_2}^{\downarrow I_1 \cap I_2}$.

In the following we consider the case of higher dimensions giving rise to non-distributive combinations. We assume that the representation, $H$, of a valuation, $V$, in a BURS is always minimal and homogeneous (all the valuations are defined in the same frame as $V$).

**Definition 17** A valuation $V' \in \mathcal{V}^*$ is said to be in a $l$-dimensional face of minimal set of basic upper valuations $H$ if and only if there is a subset $R \subseteq H$ such that $|R| = l + 1$ and $\oplus R \preceq V'$.

**Definition 18** $V' \in \mathcal{V}^*$ is said to be an extreme point of minimal set of basic upper valuations $H$ if and only if $V'$ is in a face of dimension 0.

**Definition 19** $V'$ is said to be interior to $\oplus H$ of dimension $l$, if and only if is not in a face of dimension $l' < l$ of minimal set $H$.

**Definition 20** $\otimes$ is said to preserve interior points if and only if ($V'$ interior to $(\oplus H_1)$ and to $(\oplus H_2)$) implies ($V'$ interior to $(\oplus H_1) \otimes (\oplus H_2)$).

**Definition 21** Given $H_1$ and $H_2$ set of basic upper valuations in $\mathcal{V}_I$, then the pair $(R_1, R_2)$ is said to be minimal consistent (MC) if and only if:

1. $R_1 \subseteq H_1, R_2 \subseteq H_2$.

2. $(\oplus R_1) \otimes (\oplus R_2) \neq V_c^{\downarrow I}$

3. If $(R'_1, R'_2)$ verifies 1) and 2) then $R_1 \subseteq R'_1$ and $R_2 \subseteq R'_2$.

**Theorem 5** If $\otimes$ preserves interior points, then

$$(\oplus H_1) \otimes (\oplus H_2) = \oplus\{(\oplus R_1) \otimes (\oplus R_2) : (R_1, R_2) \text{ is MC }\} \quad (38)$$

Proof

Let $(\oplus H_1) \otimes (\oplus H_2) = \oplus\{V_1, \ldots, V_m\}$ be a minimal representation of $(\oplus H_1) \otimes (\oplus H_2)$.

For each $V_i$, let $R_i^j (j = 1, 2)$ a minimal subset of $H_j$ such that $(\oplus R_i^j \preceq V_i)(j = 1, 2)$.

$V_i$ is extreme of $(\oplus R_i^1) \otimes (\oplus R_i^2)$ because it is extreme of $(\oplus H_1) \otimes (\oplus H_2)$ and

$$(\oplus H_1) \otimes (\oplus H_2) \preceq (\oplus R_i^1) \otimes (\oplus R_i^2)$$

$V_i$ *is interior to* $(\oplus R_i^1)$ *and* $(\oplus R_i^2)$, *so it is interior to* $(\oplus R_i^1) \otimes (\oplus R_i^2)$.

*As $V_i$ is interior and extreme of $(\oplus R_i^1) \otimes (\oplus R_i^2)$, then the only possibility is that $(\oplus R_i^1) \otimes (\oplus R_i^2) = \{V_i\}$. And therefore $(\oplus R_i^1) \otimes (\oplus R_i^2)$ is minimal consistent.*

Theorem above allows to reduce the number and the size of the combinations.

**Example 12** *When valuations are polyhedral in $\mathbb{R}_I$, then the dimension of a valuation $\oplus H \subseteq \mathbb{R}_I$ is $|I|$ minus the dimension of the smaller affine subspace containing $\oplus H$.*

*Of course intersection of polytopes preserves interior points and the combination can be carried out using Theorem 5.*

*The efficiency of the resulting algorithm will depend of the efficiency in the determination of minimal consistent pairs. We do not know any general algorithm based on this search. In fact, in the classical book by Preparata and Shamos (1985), this problem (intersection of convex sets given by their extreme points) is only considered in dimensions 2 and 3, and the algorithms are based on quite sophisticated techniques.*

## 5 CONCLUSIONS

We have studied idempotent valuations proposing new algorithms to work with sets and polytopes in big frames. The methods of calculating are also applicable to other idempotent valuations systems such as Possibility Theory (Dubois and Prade 1991, Shenoy 1992), or the Symbolic Theory of Evidence (Kohlas 1993).

The idea is to use to dual strategies to represent and calculate with valuations: as infimum of very informative valuations or supremum of very low informative ones. We do not think that working with basic representation systems is essential. We can use a non-basic representation system, but then more effort should be devoted to determine efficient ways of doing marginalization or combination.

A refinement of some of the resulting algorithms and comparison of them with another classical procedures is necessary in the future.

### References

J.E. Cano, M. Delgado, S. Moral (1993) An axiomatic framework for propagating uncertainty in directed acyclic graphs. *International Journal of Approximate Reasoning* 8, 253-280.

J.E. Cano, S. Moral, J.F. Verdegay-López (1993) Propagation of convex sets of Probabilities in directed acyclic networks. In: *Uncertainty in Intelligent Systems* (B. Bouchon-Meunier, L. Valverde, R.R. Yager, eds.) Elsevier, Amsterdam, 15-26.

D. Dubois, H. Prade (1991) Inference in possibilistic hypergraphs. In: *Uncertainty in Knowledge Bases* (B. Bouchon-Meunier, R.R. Yager, L.A. Zadeh, eds.) Springer-Verlag, Berlin, 250-259.

J. Kohlas (1993) Symbolic evidence, arguments, supports and valuation networks. In: *Symbolic and Quantitative Approaches to Reasoning and Uncertainty* (M. Clarke, R. Kruse, S. Moral, eds.) Springer Verlag, Berlin, 186-198.

J. Kohlas, S. Moral (1996) Propositional Information Systems. Working Paper N. 96-01, Institute of Informatics, University of Fribourg.

J. Kohlas, R.F. Stärk (1996) Information algebras and information systems. Working Paper N. 96-14, Institute of Informatics, University of Fribourg.

C. Lassez, J.L. Lassez (1992) Quantifier elimination for conjunctions of linear constraints via a convex hull algorithm. In: *Symbolic and Numerical Computation for Artificial Intelligence*, (B.R. Donald et al., eds.) Academic Press, London, 103-119.

T.H. Mattheis, D.S. Rubin (1977) A survey and comparison of methods for finding all vertices of convex polyhedral sets. Technical Report N. 77-14, Department of Operations Research and Systems Analysis, University of Carolina at Chapel Hill.

F.P. Preparata, M.I. Shamos (1985) *Computational Geometry. An Introduction.* Springer-Verlag, New York.

D.S. Scott (1982) Domains for denotational semantics. In: *Automata, Languages and Programming*, (M. Nielsen, E.M. Schmitt, eds.) Springer Verlag, 577-613.

G. Shafer (1991) An axiomatic study of computation in hypertrees. Working Paper N. 232. School of Business, University of Kansas, Lawrence.

G. Shafer, P.P. Shenoy (1988) Local Computation in Hypertrees. Working Paper N. 201. School of Business, University of Kansas, Lawrence.

P.P. Shenoy (1992) Using possibility theory in expert systems. *Fuzzy Sets and Systems* 52, 129-142.

P.P. Shenoy, G. Shafer (1990) Axioms for probability and belief-functions propagation. In: *Uncertainty in Artificial Intelligence*, *4* (R.D. Shachter, T.S. Levitt, L.N. Kanal, J.F. Lemmer, eds.) North-Holland, Amsterdam, 169-198.